\documentclass[9pt,twocolumn,twoside,final]{pnas-new}

\makeatletter
\fancypagestyle{firststyle}{
  \fancyfoot[RO]{\footerfont Preprint — October 2025 — \thepage}
  \fancyhead{}

}

\fancypagestyle{plain}{
  \fancyfoot[RO]{\footerfont \thepage — Preprint — October 2025}
  \fancyhead{}

}

\makeatother

\templatetype{pnasbriefreport}

\title{The Method of Infinite Descent}

\author[a]{Reza T. Batley}
\author[a,1]{Sourav Saha}

\affil[a]{Kevin T. Crofton Department of Ocean and Aerospace Engineering, Virginia Polytechnic Institute and State University, Blacksburg VA, United States}

\leadauthor{Batley}

\authorcontributions{R.T.B. conceived the study, developed the theoretical framework, derived the analytical formulation, implemented the numerical demonstrations and wrote the manuscript. S.S. supervised the work, and provided critical feedback and revisions.}
\authordeclaration{The authors declare no competing interests.}
\correspondingauthor{\textsuperscript{1}To whom correspondence should be addressed. E-mail: souravsaha@vt.edu}


\begin{abstract}
Training - the optimisation of complex models - is traditionally performed through small, local, iterative updates [D. E. Rumelhart, G. E. Hinton, R. J. Williams, Nature 323, 533–536 (1986)]. Approximating solutions through truncated gradients is a paradigm dating back to Cauchy [A.-L. Cauchy, Comptes Rendus Mathématique 25, 536–538 (1847)]  and Newton [I. Newton, The Method of Fluxions and Infinite Series (Henry Woodfall, London, 1736)]. This work introduces the \emph{Method of Infinite Descent}, a semi-analytic optimisation paradigm that reformulates training as the direct solution to the first-order optimality condition. By analytical resummation of its Taylor expansion, this method yields an exact, algebraic equation for the update step. Realisation of the infinite Taylor tower's cascading resummation is formally derived, and an exploitative algorithm for the direct solve step is proposed.

This principle is demonstrated with the herein-introduced AION (\emph{Analytic, Infinitely-Optimisable Network}) architecture. AION is a model designed expressly to satisfy the algebraic closure required by Infinite Descent. In a simple test problem, AION reaches the optimum in a single descent step. Together, this optimiser-model pair exemplify how analytic structure enables exact, non-iterative convergence. Infinite Descent extends beyond this example, applying to any appropriately closed architecture. This suggests a new class of semi-analytically optimisable models: the \emph{Infinity Class}; sufficient conditions for class membership are discussed. This offers a pathway toward non-iterative learning.
\end{abstract}

\begin{document}

\maketitle
\thispagestyle{firststyle}
\ifthenelse{\boolean{shortarticle}}{\ifthenelse{\boolean{singlecolumn}}{\abscontentformatted}{\abscontent}}{}

\firstpage[3]{5}

\dropcap{N}onlinear optimisation is fundamental to science. It is well established that such algorithms proceed by means of small, finite, local steps. This paradigm, traceable to the early works of Cauchy \cite{cauchy1847methode} and Newton \cite{newton1736fluxions}, has long served as the bedrock of optimisation. This paradigm - this limitation - however, may not be inherent to optimisation itself; rather a consequence of the arbitrary structure besetting the models we seek to optimise. This raises a natural question: is there a universal function approximator whose structure renders its optimality condition analytically exact? 

The answer lies in a novel class of neural models: separable neural architectures (SNAs). This is a family of structured models with learnable basis functions, \emph{atoms} whose interactions are not arbitrary but instead engineered. Models encompassed in the SNA family include the Interpolating Neural Network \cite{park2025unifying}, the Separable Physics-Informed Neural Network \cite{cho2023separable} and KHRONOS \cite{batley2025khronos}. This unified framework allows for the creation of models ranging from simple additive forms to tensor decompositions. This present work focuses on a subclass in which functions are represented as a sum of products of univariate bases. Within this structure, one identifies the \emph{Infinity Class}. This SNA subclass is defined as those structured models whose optimality condition remain analytically closed under Taylor expansion; any high-order term is expressible in finite algebraic form. This is, in effect, algebraic closure over differentiation and multiplicative composition.

A particular member of this class, introduced herein, is AION (\emph{Analytic, Infinitely-Optimisable Network}). In AION, each univariate basis is a dense linear combination of exponential-trigonometric functions. This is crucial for two reasons: first, this guarantees that each basis is dense in the space of continuous functions, a necessary condition for the model's universal approximation capability. Second, the exponential-trigonometric form endows the model with that key algebraic closure. In fact, this allows for analytic resummation of the Taylor series of its optimality condition. As shown below, this closure has implications for the structure and solvability of the optimisation problem.

\section*{Results}
This section opens by establishing, in precise form, the centrepiece of the present study: AION. The form of this model is chosen to satisfy the desired properties of universality and resummation of its Taylor expansion. Situated in the Euclidean space of dimension $d$ with coordinates $x=(x_1,\dots,x_d)$, and for a given rank $r$ - that is, the total number of separable terms - this may be defined as the function $f^{\infty}_r:\mathbb{R}^d\rightarrow\mathbb{R}$ of the form,
\begin{align}
    f^{\infty}_r(x;\Theta)=\sum_{j=1}^r\prod_{i=1}^d \psi_i^{(j)}\left(x_i;\theta_i^{(j)}\right),
\end{align}
\emph{atoms}, $\psi_i^{(j)}=\sum_{p=1}^PA^{(j)}_{ip}\exp{\left(\alpha^{(j)}_{ip}x_i+i\left(\omega^{(j)}_{ip}x_i+\varphi^{(j)}_{ip}\right)\right)}.$
The $i$-th atom of the $j$-th rank has the learnable parameter set $\theta^{(j)}_i=\{A^{(j)}_{ip},\alpha^{(j)}_{ip},\omega^{(j)}_{ip},\varphi^{(j)}_{ip} \in \mathbb{R}\}_{p=1}^P$ with the standard wavelet-style parameters of \emph{amplitude} $A$, localised \emph{growth/decay} $\alpha$, \emph{frequency} $\omega$ and \emph{phase} $\varphi$. The total learnable parameter dictionary is then $\Theta=\{\theta^{(j)}_i\}_{i,j}$. Henceforth, the atomic exponent will be compacted to something akin to a dot product $\langle a^{(j)}_{ip},x_i\rangle$ with vector $a=(\alpha,i\omega,i\varphi)^T\in\mathbb{R}\times(i\mathbb{R})^2$ capturing the core parameters.

Indeed, an immediate property that follows is the density of atoms in $C(\mathbb{R})$, familiar from Fourier-analytic tradition. From this it follows that the full architecture $f^{\infty}_r$ inherits density in $C(\mathbb{R^d})$, by a direct application of the Stone-Weierstrass theorem \cite{stone1948weierstrass}. The next is rooted in the first-order optimality condition of the loss functional $\Phi(\Theta)$ by which the architecture is trained. In the canonical setting of paired data $\{x^{(n)},y^{(n)}\}_{n=1}^N$ one typically arrives at the least-squares objective, $\Phi(\Theta)=\sum_{n=1}^N\|f^{\infty}_r(x^{(n)};\Theta)-y^{(n)}\|^2$. For clarity of exposition, further analysis shall proceed with this loss.

To this end, recall the \emph{first-order optimality condition}: at any local minimum $\Theta^*$ of the objective, the gradient must vanish - $\nabla\Phi(\Theta^*)=0$. Starting from some point $\Theta$, the \emph{training} of an architecture can simply be cast as finding that $\Delta$ for which $\nabla\Phi(\Theta+\Delta)=0$. Differentiating yields 
\begin{align}
    \sum_{n=1}^N2\left(f^\infty_r(x^{(n)};\Theta+\Delta)-y^{(n)}\right)\nabla_{\Theta}f^{\infty}_r(x^{(n)};\Theta+\Delta).
\end{align}
Herein lies the crux: in any other conventional setting one must resort to approximation of the $\nabla_{\Theta}f^{\infty}_r(x^{(n)};\Theta+\Delta)$ term by invoking and truncating its Taylor expansion. This reduction to locality - the tentative, approximate steps often assumed inherent - arises not from the optimisation itself, but from the architecture upon which it is enacted. Indeed, the exponential structure of AION is fashioned in such a sense that it remains closed under both differentiation and multiplication.

This closure is decisive; a conventional architecture computing the gradient at the shifted parameter point $\Theta+\Delta$ inextricably traps the update $\Delta$ within non-analytic functions. The closure of Infinity-Class SNAs, however, allows $\Delta$ to cleanly factorise out into its own multiplier. Proceeding, it is notable then that each differentiation of an atom $\psi^{(j)}_i$ yields i) polynomial prefactors; and ii) retention of exponential-trigonometric form. Each component of $\nabla\Phi(\Theta)$ may be written as the finite sum $\sum_{\ell\in\mathcal{L}}P_\ell(\Theta)B_\ell(\Theta)$, with $B$ the basis factor
\begin{align}
    B^{(j,n)}_{ip}(\Theta)=e^{\langle a^{(j)}_{ip}, x^{(n)}_i\rangle}\prod_{s\neq i}\psi^{(j)}_{s}(x^{(n)}_s; \theta^{(j)}_s).
\end{align}
For each $\ell$, each parameter shift acts termwise. Noting that 
\begin{align}
    e^{\Delta\cdot\nabla_\Theta}[P_\ell(\Theta)B_\ell(\Theta)]=P_\ell(\Theta+\Delta)e^{\langle a_\ell, \Delta\rangle}B_\ell(\Theta),
\end{align}
and the classical result due to analyticity of $\Phi$, $\nabla\Phi(\Theta+\Delta)=\exp({\Delta\cdot\nabla_\Theta})[\nabla\Phi(\Theta)]$ \cite{olver2014introduction}, one obtains
\begin{align}
    \nabla\Phi(\Theta+\Delta)=\sum_{\ell\in\mathcal{L}}P_\ell(\Theta+\Delta)e^{\langle a_\ell, \Delta\rangle}B_\ell(\Theta).
\end{align}
This is the desired analytic resummation. It is here the rootfinding nature of the problem becomes manifest. Writing $b(\cdot)=B_\ell(\cdot),p(\cdot)=P_\ell(\cdot)$ and $D(\cdot)=\operatorname{diag}(e^{\langle a_\ell,\cdot \rangle})$ one arrives at the structured, nonlinear system
\begin{align}
    p(\Theta+\Delta)^TD(\Delta)b(\Theta)=0.
\end{align}
Calculating the update $\Delta$ thus reduces to solving this set of equations. Its cardinality is formally $|\mathcal{L}|=NrdP$, yet the dependence on $N$ is merely in aggregation. Indeed,
\begin{align}
    \label{eq:f}
    \sum_{n=1}^N p^{(n)}(\Theta+\Delta)^TD(\Delta)b^{(n)}(\Theta)=0.
\end{align}
The effective dimensionality is then $O(rdP)$, no different than stochastic gradient descent. For convience this problem will be abbreviated to $F(\Delta)=0$.
\subsection*{Algorithmic Realisation: Structured Newton Raphson (SNR)}
Black-box rootfinding typically proceeds by finite differencing of $F$, but its neat structure allows its analytical precomputation. Note that whilst the ensuing algorithm involves \emph{inner} iterations, these should in no way be mistaken for the incrementalism of conventional truncated-gradient approaches. In Infinite Descent, iterations do not occur in optimisation space - on the loss landscape; rather, they occur within the exact algebraic root of the analytically resummed system. It does not approximate a trajectory; it resolves a closed-form equation whose root corresponds to the "one-shot" update step. These inner iterations merely \emph{expose}, not approximate, this step.

Consider the $n$-th component of $J(\Delta)=\partial_\Delta F$, as the derivative and summation orders can be freely interchanged. The product of $p^T$ and $D$ as in \eqref{eq:f} ensures the Jacobian yields two terms. The second is simply $p^{(n)}(\Theta+\Delta)^T\operatorname{diag}(a_\ell e^{\langle a_\ell,\Delta \rangle})b(\Theta)$, self-evidently block-diagonal, even diagonal in each rank. The first term's block-diagonal nature follows from the model's separability. Indeed, each $p^{(n)}$ depends only on the parameters of its own rank, derivatives with respect to any other annihilating that dependence; any $j\neq j'$ leads to $\partial_j\partial_{j'}f^\infty_r=0$. All cross-partials thereby vanish, cleanly decomposing the Jacobian to $J(\Delta)=\operatorname{diag}(J_1(\theta^{(1)}),\dots J_r(\theta^{(r)}))$. 

In fact, each $J_j$ is not dense; its structure enables a Kronecker product representation: $J_j=J_{j,1}\otimes\cdots\otimes J_{j,d}$. Plainly, differentiation proceeds in isolation along orthogonal directions: for each rank its own stream, for each dimension its own course. This collapses the dimensional cost from cubic to linear: an absolute worst-case of $O(rdP^3)$ for a dense Newton-Raphson iterate. This can be compressed further by accounting for the low-rank nature of each atomic block, so even this is pessimistic. Nevertheless, the examples that follow shall employ the full Newton-Raphson formulation. To this end, for the $j$-th block, the iterate is $\delta_j=-J_j^{-1}F_j(\Delta)$, with damping, line-search or other stabilisation applied if required. To illustrate the practical consequences of this structure, a simple numerical experiment is presented below.

\subsection*{Demonstration}
A simple toy problem illustrates the method. This problem is prescribed by the function $g(x,y)=\cos(\pi(x-y))$, sampled uniformly on a $25\times 25$ grid over $(x,y)\in[0,1]^2$. A rank $r=2$, $P=1$ ICNSA is initialised, totalling eight trainable parameters, each rank having its own amplitude $A$, growth/decay $\alpha$, frequency $\omega$ and phase $\varphi$. As $g$ can be written as a simple sum of a product of trigonometric functions, this setup is sufficiently expressive to approximate it to effectively analytic precision.

\begin{figure}[h!]
\centering
\includegraphics[width=0.85\linewidth]{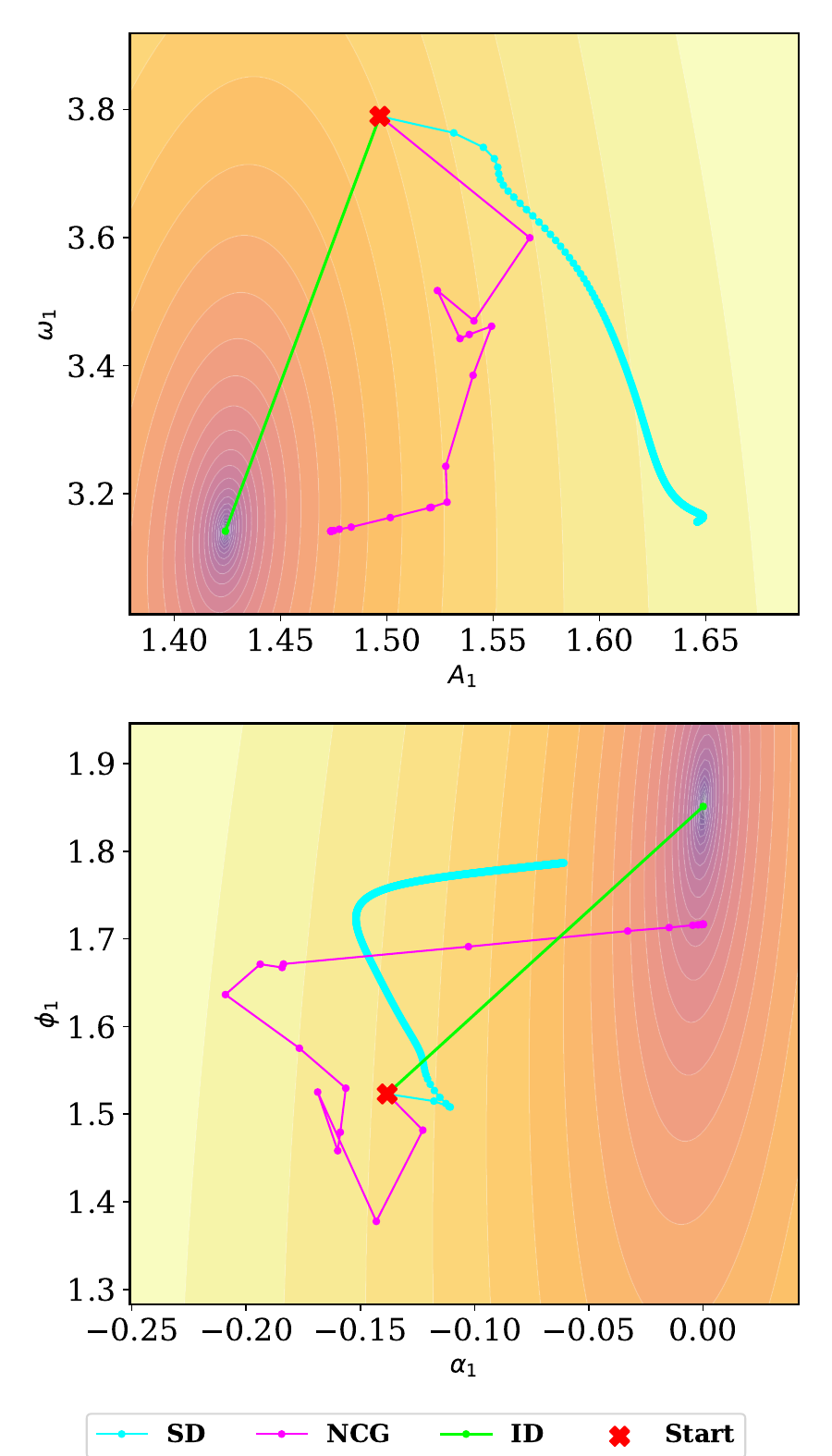}
\caption{Level sets $\alpha_1\times\phi_1$ and $A_1\times \omega_1$ of the loss landscape. The Method of Infinite Descent (ID; lime), leaps from the initial point (red cross) to the minimum. The Method of Steepest Descent (SD; cyan) and Newton Conjugate Gradient (NCG; magenta) follow slower, locally-informed paths.}
\label{fig:fig1}
\end{figure}

The Method of Infinite Descent (ID) - initialisation $\rightarrow$ resummation $\rightarrow$ SNR - is compared against two canonical optimisers: Steepest Descent (SD) and Newton Conjugate Gradient (NCG) \cite{nocedal2006numerical}. SD is implemented with Armijo backtracking line search for 1,000 iterations, with Armijo constant $c_1=1\times 10^{-4}$, step factor $\rho=0.5$ and unit initial step length. The NCG implementation is that taken from \texttt{scipy.optimize.minimize}, running either for 50 iterations or until the loss dips below $10^{-8}$, whichever occurs first.

As summarised in Table \ref{tab:tab1}, the Method of Infinite Descent attains the minumum in a single leap, notwithstanding inner rootfinding iterations, based on its analytically resummed infinite-gradient information. Both canonical optimisers trace incremental paths guided by locally-truncated gradient steps.  The visualisation in Figure \ref{fig:fig1} illustrates this contrast in a fashion more stark. This empirical behaviour affirms that the apparent approximation locality of conventional methods arises not from the principles themselves, but from the analytic incompleteness of the models upon which they act.

\begin{table}[h!]
\centering
\caption{A comparison of Infinite Descent (ID), Steepest Descent (SD) and Newton Conjugate Gradient (NCG) on a toy problem}
\begin{tabular}{lccc}
\label{tab:tab1}
Method & Iterations & Walltime (ms) & Final Loss \\
\midrule
SD & 1000 & 6123 & $6\times 10^{-6}$ \\
NCG & 28 & 383 & $5\times 10^{-13}$ \\
ID & 1 & 102 & $9\times 10^{-18}$\\
\bottomrule
\end{tabular}
\end{table}
\section*{Discussion}
This work introduced the Method of Infinite Descent, a semi-analytic approach to exact optimisation in structured models. It reformulates training as the direct solution of the first-order optimality condition, enabled here through the analytic closure of the underlying architecture. Demonstration of this principle was done with the proposed Analytic, Infinitely-Optimisable Network (AION). AION is proposed as an instance of \emph{Infinity-Class} Separable Neural Architectures, a family of models of structure permits such analytic treatment. Indeed, AION exemplifies this structure and demonstrates this method by "one-shotting" a toy problem. 

It is, however, important to note that the broader contribution of this work is in the Method of Infinite Descent itself. Beyond AION, the \emph{Infinite Descent} framework lends itself to any architecture exhibiting algebraic closure. This suggests swathes of unexplored ideas. Future work will systematically explore this space: identifying candidate architectures, generalising the resummation principle and testing the limits of non-truncated optimisation in higher-dimensional and stochastic settings. Furthermore, while the present implementation required inner iterations to solve the resultant rootfinding system, future work might focus on its reduction or replacement. This may entail symbolic factorisation of the root structure, or developing partial-closure architectures with an invertible inner solve. Perhaps this work could lay the groundwork for a powerful and applicable new model.

\bibsplit[4]

\bibliography{pnas-sample}

\end{document}